\documentclass[conference]{IEEEtran}
\IEEEoverridecommandlockouts
\usepackage{cite}
\usepackage{amsmath,amssymb,amsfonts}
\usepackage{algorithmic}
\usepackage{graphicx}
\usepackage{textcomp}
\usepackage{xcolor}
\usepackage{multirow}
\usepackage{hyperref}

\newcommand{\bs}[1]{{\boldsymbol{#1}}}
\def\BibTeX{{\rm B\kern-.05em{\sc i\kern-.025em b}\kern-.08em
    T\kern-.1667em\lower.7ex\hbox{E}\kern-.125emX}}
\begin{document}

\title{Categorical EHR Imputation with Generative Adversarial Nets
}

\author{\IEEEauthorblockN{1\textsuperscript{st} Yinchong Yang}
\IEEEauthorblockA{\textit{Siemens AG} \\
Munich, Germany \\
yinchong.yang@siemens.com}
\and
\IEEEauthorblockN{2\textsuperscript{nd} Zhiliang Wu}
\IEEEauthorblockA{\textit{Siemens AG} \\
\textit{Ludwig-Maximilians University of Munich}\\
Munich, Germany \\
zhiliang.wu@siemens.com}
\and
\IEEEauthorblockN{2\textsuperscript{nd} Volker Tresp}
\IEEEauthorblockA{\textit{Siemens AG} \\
\textit{Ludwig-Maximilians University of Munich}\\
Munich, Germany \\
volker.tresp@siemens.com}
\and
\hspace{6cm}\IEEEauthorblockN{3\textsuperscript{rd} Peter A. Fasching}
\IEEEauthorblockA{\hspace{6cm}\textit{University Clinics Erlangen} \\
\hspace{6cm}\textit{Department of Gynecology and Obstetrics}\\
\hspace{6cm}Erlangen Germany \\
\hspace{6cm}peter.fasching@uk-erlangen.de}
}

\maketitle

\begin{abstract}
Electronic Health Records often suffer from missing data, which poses a major problem in clinical practice and clinical studies. A novel approach for dealing with missing data are Generative Adversarial Nets (GANs), which have been generating huge research interest in image generation and transformation.
Recently, researchers have attempted to apply GANs to missing data generation and imputation for EHR data: a major challenge here is the categorical nature of the data.
State-of-the-art solutions to the GAN-based generation of categorical data involve either reinforcement learning, or learning a bidirectional mapping between the categorical and the a real latent feature space, so that the GANs only need to generate real-valued features.
However, these methods are designed to generate complete feature vectors instead of imputing only the subsets of missing features.
In this paper we propose a simple and yet effective approach that is based on previous work on GANs for data imputation.
We first motivate our solution by discussing the reason why adversarial training often fails in case of categorical features.
Then we derive a novel way to re-code the categorical features to stabilize the adversarial training.
Based on experiments on two real-world EHR data with multiple settings, we show that our imputation approach largely improves the prediction accuracy, compared to more traditional data imputation approaches. 
\end{abstract}


\section{Introduction}
\iffalse
Story:
\begin{itemize}
    \item[1]: importance of data quality
    \item[2]: multiple instead of single imputation
    \item[3]: GAN as suitable solution (with certain modification)
    \item[4]: new challenge in EHR: categorical features
    \item[5]: we propose...
    \item[6]: overview of the paper
\end{itemize}
\fi

\textbf{The increasing importance of data quality in healthcare:} 
Electronic Health Records (EHR) present a rich data source and are, e.g., used for intra- and inter-departmental information exchange, for documentation purposes, and, most recently,  as the basis for many analytic studies.
Typically, data involving critical clinical decision paths are of good quality, but less critical data are often incomplete, e.g., due the huge workload of clinical personnel; this poses a significant problem for the secondary use of EHR data.
In particular the value of  a clinical study greatly depends on data completeness and correctness.
Although the prime solution would be to enhance the  EHR quality by improving the EHR system design and the  data collection process,
the missing data problem is not likely to completely disappear. \cite{little2014statistical} provides an overview on missing data approaches in statistics and
\cite{tresp1995efficient} presents solutions to the neural network setting.
When using nonlinear models, data imputation is often used \cite{wells2013strategies,mirkes2016handling}, which is also the approach pursued in this paper.
Data imputation is often based on parametric or nonparametric probability density estimation.
In this paper we investigate a recently developed GAN architecture. It imputes data without calculating a probability density first, and might become an important  method of choice in the future. 

\textbf{Multiple instead of single imputation:}
More specifically, we discuss a novel realization of the well-known multiple imputation approach \cite{stuart2009multiple, little2014statistical, rubin2004multiple}.
By embedding certain randomness into the imputation method and performing imputation multiple times, one can achieve more flexibility and reliability than with single imputation.
This allows for ---in contrast to an averaged point estimate of each missing value--- estimating the statistical reliability of the imputation methods \cite{donders2006gentle}.
Multiple Imputation by Chained Equations (MICE) \cite{raghunathan2001multivariate} fits one regression model for each feature that contains missing values, conditioned on all complete features.
This method can model the dependency between features but the number of necessary regression models increases quadratically with the number of features.
One could also simply assume a multivariate Gaussian distribution for the missing features and draw multiple random samples as imputation.
The covariance matrix represents the dependency between features but the Gaussian distribution cannot handle categorical features, which are often present in EHR data.
In this paper we investigate a novel approach that takes into account the categorical nature of the features while modeling inter-feature dependency in an efficient way.
%\textsc{Comment: you describe the \cite{raghunathan2001multivariate} approach in great detail. You should mention why (is our approach an extension of that one?)}
%Comment: I have reduced the description of this method, in order to avoid misunderstanding.

\textbf{GAN as multiple imputation:}
In recent years, a new class of neural networks, called Generative Adversarial Nets (GANs), have been developed and have generated huge interest in the research community.
The original paper \cite{goodfellow2014generative} proposes to train a network that can learn the underlying distribution of the data, allowing for generating unlimited amount of  data instances. When applied to images, the generated images often appear quite real.
Since their initial introduction, a large variety of exciting improvements and modifications of the GAN framework have been proposed to solve different and yet related tasks, such as generating labeled data \cite{mirza2014conditional}, image translation \cite{CycleGAN2017}, deriving super-resolution \cite{ledig2017photo} and image augmentation \cite{shrivastava2017learning}.
\cite{yoon2018gain} proposes a new variant, the Generative Adversarial Imputation Nets (GAINs), to perform data imputation and shows promising results on multiple benchmark datasets.
This method presents in fact a novel realization of the multiple imputation concept, and it is also related to the MICE algorithm.
Instead of trying all possible orders to build the regression chain, it exploits the expressiveness of deep neural networks to model all features with missing values simultaneously.
However, this approach cannot immediately be applied to EHR data, as we discuss now.

\textbf{Challenge in EHR data for GAN:}
Most of the GAN models have been designed for image data, where the features, i.e., the pixel values, are real numbers.
This enables error back-propagation within the GAN framework.
In EHR data, however, a large proportion of patient features are categorical.
In order to generate categorical data, even when binary coded, requires operations that are not differentiable, meaning that standard adversarial training is not possible.
Due to the same reason, GANs have not seen many successful applications in NLP data \cite{rajeswar2017adversarial}. %\textsc{(can you cite something here?)} Comment: yes, although it's just some pre-mature overview. My actual reference is a post by goodfellow on this question at reddit.
A few approaches to generate discrete data with GANs have recently been proposed; most promising are approaches involving reinforcement learning \cite{yu2017seqgan}, more specifically policy gradients \cite{li2017adversarial}.
Another proposed solution is to learn a mapping function from the discrete space of words to a latent real space as well as a reverse mapping \cite{zhang2017adversarial}.
\cite{choi2017generating} applies this idea to handle categorical features in EHR data and develops auto-encoders to function as the mapping.
In such cases, the GAN model only needs to generate real valued vectors that represent the originally discrete data instances, allowing for the gradient propagation from discriminator and generator.
In our related works section, we will review this approach in more detail.
It is to note that the mapping functions between discrete and real spaces serve as pre- and post-processing steps, and are crucial for the quality of the translation between the discrete and real spaces.
Such mapping functions are often trained in an Auto-encoder fashion and thus rely on the completeness of the input features.
In a data imputation setting, however, the input features are by definition incomplete and the learned mappings must learn to map incomplete data.
That is to say, this proposed approach is only applicable to generating complete feature vector instead of subsets of features.

\textbf{Our contributions:} The Generative Adversarial Imputation Nets framework \cite{yoon2018gain} has been proposed to apply adversarial training to impute missing data of real values.
In this work, we adjust this framework so that it can also perform data imputation for categorical features.
We hypothesize that the reason that adversarial training often fails with softmax activation in the generator is that, while the true data features contain exclusively 0s and 1s, the softmax function can only produce a probability value between 0 and 1.
On one hand, within a couple of epochs, the discriminator with sufficient expressiveness can learn to discriminate the generated values from real data exploiting this fact. On the other hand, it typically takes more epochs of training before the generator can produce real values close to 0 and 1.
This phenomenon, i.e., that the discriminator always makes correct decisions and the generator always receives negative feedback from the very beginning, results in the divergence of the adversarial training \cite{arjovsky2017wasserstein}.
In other words, the generator fails to learn anything useful to improve itself. %(comment: from the logic of the paper this part should come earlier!)

One of our major contributions is to propose a small but very effective modification to the data processing step.
We perform a fuzzy binary coding of categorical features, i.e., we encode the binary values using real numbers between 0 and 1, while retaining the category information.
In this way we guarantee that from the very beginning of the adversarial training, values produced by the generator already resemble the real values in their domain.
To this end, the discriminator can not ``cheat'' and exploit the simple fact that real data are all binary while generated data are all real.
The discriminator can only focus on the true and informative characteristics of the real and generated data, such as the dependency between features.
Thus, the generator can receive more useful gradient updates from the discriminator, which improves the data generation process.

%The GAIN framework can be seen as a development of the conditional GAN to generate labeled data instances.
%Analogously to the label information, there is a mask vector, encoding which features are missing, that serves as a condition to the generator and discriminator.
%The latter one then attempts to predict the mask vector.
%Our major contribution in adjusting this framework is twofold:
%First we show that one needs specific discriminators that assign one probability t
%Second and most importantly
%We also demonstrate that the imputed data improve predictive modeling based on realworld EHR datasets.
The rest of this paper is organized as follows.
In section \ref{sec:related} we provide an overview of related works in three research fields of GANs: generation of categorical data, application GANs for EHR data and for data imputation.
After a brief introduction to the GAN framework in section \ref{sec:preliminary}, we elaborate the methods we propose in detail, including the fuzzy binary coding and the GAN for categorical data generation in section \ref{sec:method}.
In section \ref{sec:experiments} we present our experimental results on two EHR datasets and show that imputation based on the GAN framework with fuzzy binary coding can be quite effective in dealing with missing categorical data in EHRs.

\section{Related Works} \label{sec:related}

%\textsc{comment: you explain details without introducing the GAN model, so maye this should come later.}
% I'm afraid I do not have time to figure out an intro to GAN that is so concise to fit in here... Also, I expected GAN to be textbook materials by now anyway.

\textbf{GANs that generate categorical features:}
There are currently three different approaches for generating categorical features with GANs.
The first approach modifies the output activation function in the generator so that the gradient can flow from the discriminator to the generator while the latter generates pseudo-discrete features. Examples are so-called Gumbel Softmax \cite{kusner2016gans, jang2016categorical} or a soft argmax function \cite{zhang2016generating}.
The second approach modifies the training objectives. \cite{yu2017seqgan, li2017adversarial} apply REINFORCE \cite{williams1992simple} algorithm for adversarial training.
The third approach, including \cite{zhang2017adversarial, choi2017generating}, learns a mapping from the raw discrete feature space to a latent real space, as well as the reverse mapping.
These mapping functions are, e.g., realized as an auto-encoder.
With the first mapping one transforms all training data that are originally categorical into real representations.
Then, the GANs framework only has to operate in this real space, learning to generate real feature vectors.
As a post processing step, the generated vectors are transformed back into the discrete space using the second mapping function.

\textbf{GANs in EHR data analysis:}
GANs have already found various interesting applications in healthcare.
\cite{choi2017generating} and \cite{esteban2017real} aim at generating pseudo-synthetic EHR data for the purpose of de-identification. The former focuses on the challenge of generating categorical features by applying an auto-encoder that can map between the discrete feature space and a real latent space. It is pointed out that applying differentiable Gumbel softmax or soft argmax functions does not completely solve the categorical problem, because patient features could be multinomial (i.e., multiclass)
as well as multiple Bernoulli distributed (i.e., multi-label).
The latter paper develops GANs that are based on Recurrent Neural Networks (RNN) to generate high dimensional time series EHR data.

\textbf{Missing Data Imputation using Generative Adversarial Nets:} \cite{yoon2018gain} adjusted the GAN framework for the specific task of data imputation.
It can be interpreted as a special case of conditional GAN, in the sense that both discriminator and generator take as input a mask vector indicating the missingness of feature values.
It is shown that this novel training framework can efficiently impute real-valued features, especially in case where the missing rate is relatively high.
Our method is largely inspired by this work, but we focus on the specific techniques to perform adversarial imputation of categorical features.

\section{Preliminary: The Generative Adversarial Nets Framework} \label{sec:preliminary}
In its simplest case, a GAN framework \cite{goodfellow2014generative} consists of two neural networks.
The first one is often referred to as the \emph{generator}, which consumes as input some random seeds $\bs{r}$ and generate data instances $\bs{g}$ that are supposed to resemble real data $\bs{x}$.
The generator can be seen as a function of
\begin{align}
	\bs{g} = G(\bs{r} | \Theta_G).
\end{align}
Each generated sample is provided to the second neural network, the \emph{discriminator}, i.e., $D(\bs{g} | \Theta_D)$.
The discriminator also consumes as input the real data samples as $D(\bs{x} | \Theta_D)$.
The training of the generator and the discriminator is adversarial, in that, while the discriminator is trained to correctly classify a sample to be either real or generated, the generator learns to fool the discriminator so that it classify generated samples to be real.
More specifically, in term of the log-loss function $\mathcal{H} (a, b) = b \cdot \log(a) + (1-b) \cdot \log(1-a)$, we can write the discriminator loss and the generator loss as
\begin{align}
\begin{split}
	loss_D = &-\mathbb{E}_{\bs{x} \sim \mathcal{P}_{\text{real}}} \mathcal{H}(D(\bs{x} | \Theta_D), 1) \\
			&-\mathbb{E}_{\bs{r} \sim \mathcal{P}_{\text{seed}}} \mathcal{H}(D(\bs{g} | \Theta_D), 0)\\
		= &-\mathbb{E}_{\bs{x} \sim \mathcal{P}_{\text{real}}} \log(D(\bs{x}|\Theta_D)) \\
			&- \mathbb{E}_{\bs{r} \sim \mathcal{P}_{\text{seed}}} \log(1-D(\bs{g} | \Theta_D))\\
	loss_G = &-\mathbb{E}_{\bs{r} \sim \mathcal{P}_{\text{seed}}} \mathcal{H}(D(\bs{g} | \Theta_D), 1) \\
		= &-\mathbb{E}_{\bs{r} \sim \mathcal{P}_{\text{seed}}} \log (D(\bs{g}|\Theta_D))
\end{split} \label{eq:preliminary_gan}
\end{align}
respectively.
With sufficient training, $D$ will not be able to differentiate between real and generated samples by assigning neutral values to both cases.
$G$ will learn to map random seeds from an arbitrary distribution $\mathcal{P}_{\text{seed}}$ to the underlying distribution of the real data $\mathcal{P}_{\text{real}}$.
Denoted as $\hat{\mathcal{P}}_{\text{real}}$, this estimate of the real data distribution allows for unlimited sampling.

\section{Method} \label{sec:method}
In this section we give a detailed introduction to our method, which consists of two major components: the fuzzy binary coding and a modification of the Generative Adversarial Imputation Nets \cite{yoon2018gain} for categorical feature generation. 

\subsection{Fuzzy binary coding} \label{subsec:fuzzy_coding}
It is important to distinguish between \emph{multinomial} and \emph{multi-Bernoulli} distributed categorical features.
In the former case, the random variable is realized by taking only one single category, i.e., the categories are mutually exclusive. For instance, the estrogen-receptor status of a patient could only be either positive, negative or unknown.
In machine learning, especially if such features appear as targets, they are often referred to as \emph{multiclass} features and modelled with the softmax function.
In the \emph{multi-bernoulli} case, a categorical feature can realize more than one categories, such as the location of metastasis, which could be multiple organs at the same time, or multiple (serious) adverse events (AE/SAE) could be triggered by certain treatment. %\textsc{(comment: I would simply say that the latter case describes mutliple bnary features(?))} Comment: I tought about that but calling it binary might cause confusion with the "binary" in binary coding.
For such a feature with non-mutual exclusive categories, one often use the term \emph{multilabel}.
For a concise terminology, we adopt the convention from machine learning and refer to these two cases as multiclass and multilabel features for the rest of the paper.

Assume that we observe $p$ categorical features on one data instance and the $j$-th feature is a multiclass one, denoted as
\begin{align}
    \xi_j \in \Omega_j ~\text{where}~ |\Omega_j| = q_j
\end{align}

%\textsc{comment: why would a multi-label situation be a single feature? Maybe ok, but for me a little strange}

As the first step, we perform regular binary coding $\xi_j \rightarrow \bs{z}_j \in \{0, 1\}^{q_j}$.
We use the term \emph{inactive category} to refer to a category that is represented by 0; and an \emph{active category} is represented by a 1.
It is easy to see that the sum of all elements in $\bs{z}_j$ is strictly 1 if $\xi_j$ is of multiclass, and could be $\bs{\mathbb{N}}_0$ if $\xi_j$ is a multilabel feature.
These binary codings are also known as one-hot and multi-hot encodings, respectively.

In the second step, we transform the binary coded variable $\bs{z}_j$ in its fuzzy representation.

\paragraph{Multiclass case}  
We propose a transformation denoted as $f(\cdot)$ of $\bs{z}_j$ as:
\begin{align}
    \bs{x}_j(k) = f(\bs{z}_j(k)) = \begin{cases}
        \mathcal{U}[0, \frac{1}{q_j})       ~&\forall k: \bs{z}_j(k) = 0, \\
        1-\sum_{k}\bs{x}_j(k)               ~&\text{for}~ k: \bs{z}_j(k) = 1.
    \end{cases} \label{eq:fuzzy_multiclass}
\end{align}

%\textsc{Comment: I assume you mean that in the first case you draw a random number from the U-distribution; then you should change the notation and state this clearly;} 

Please note that we use $\bs{x}_j(k)$ to denote the $k$-th element in the vector $\bs{x}_j$, in order to avoid double subscripts; $\bs{\mathcal{U}}[a, b)$ denotes a continuous uniform distribution in the interval of $[a, b)$. Assuming any active category to be $k^*$, then each of the $q_j -1$ inactive categories is represented by a fraction $\bs{x}_j(k)$  which is uniformly sampled from $[0, \frac{1}{q_j})$. With this smoothing, we can retain exactly the same information encoded in $\bs{z}_j$. It is easy to see that,
\begin{align}
    1-\sum_{\forall k \neq k^*} \bs{x}_j(k) > \frac{1}{q_j}. \label{eq:fuzzy_ineq}
\end{align}
In other words, the left side of the inequation \eqref{eq:fuzzy_ineq}, which represents the active category $k^*$, is guaranteed to be larger than any fraction encoding an inactive category.
Operations such as $max$, $min$, $argmax$ and $argmin$ applied on the fuzzy $\bs{x}_j$ are always able to decode the same information in $\bs{z}_j$.
%\textsc{Question: do you really sample from $\mathcal{U}[0, \frac{1}{q_j})$ ? If yes, you should say that! Only once. or at each iteration ?}
%Comment: Yes I really draw samples. And in a few lines I mention that one could either do it once as pre-processing, or everytime before training epoch.
\paragraph{Multilabel case}
%$\xi_j$ consists of multiple bernoulli distributed variables, i.e., it can take more than one category and the binary coded vector $\bs{z}_j$ contains more than one 1.
Since the categories are no more mutual exclusive, we can derive a fuzzy binary coding by simply taking 0.5 instead of $\frac{1}{q_j}$ as the upper bound of uniform sampling:
\begin{align}
    \bs{x}_j(k) = f(\bs{z}_j(k)) = \begin{cases}
        \mathcal{U}[0, 0.5)~\text{for}~ \bs{z}_j(k) = 0, \\
        \mathcal{U}[0.5, 1] ~\text{for}~ \bs{z}_j(k) = 1.
    \end{cases} \label{eq:fuzzy_multilabel}
\end{align}
It is also guaranteed that the category information in $\bs{z}_j$ remains intact, since we can always recover $\bs{z}_j$ applying $I(\bs{x}_j \geq 0.5)$, where $I(\cdot)$ denotes the indicator function.

Transforming the binary codes into fuzzy binary codes prevents the discriminator from exploiting the fact that the generated values are all fractions and the real values only contain 0's and 1's.
This fuzzy binary coding, especially the samplings in Eq. \eqref{eq:fuzzy_multiclass} and \eqref{eq:fuzzy_multilabel}, can be performed only once as pre-processing step, or alternatively, prior to each training epoch.
In our experiments, we implement the first variant.

In Fig. \ref{fig:GLoss} we provide some empirical results based on our experiments, demonstrating that without the fuzzy binary coding trick, the adversarial training tends to diverge, i.e., the discriminator keeps improving itself by exploiting the obvious difference between the generated and real data.
The generator, therefore, receives no gradients from the discriminator for improvement.

Applying the fuzzy encoding, the discriminator can be forced to focus on discovering the true difference between the real and generated data in term of their distributions and dependencies instead of their different domains.
These discoveries in turn shall encourage the generator to approximate the real data distribution.

Lastly, we concatenate the feature vectors of all categorical features as
\begin{align}
    \bs{\bar{x}} = [\bs{x}_1, \bs{x}_2, ..., \bs{x}_p] =  [\bs{x}_j]_{j=1}^{p} \in [0, 1]^{\sum_{j=1}^p q_j},
\end{align}
which form the inputs to the generative adversarial imputation network.
Note here that we do not use another subscript denoting the data instance in $\bs{x}$, and simply assume that they are all i.i.d. samples.

\subsection{Categorical Generative Adversarial Imputation Nets (Categorical GAINs)}
\paragraph{Data notation}
In order to represent the missingness of data, \cite{yoon2018gain} introduced a binary mask vector $\bs{m}$ indicating which features are missing in a data instance represented by a real vector $\bs{\xi}$.
Here $\bs{m}$ and $\bs{\xi}$ have exactly the same size and each element $\bs{m}(k)$ is 1 if $\bs{\xi}(k)$ is not missing, and 0 otherwise.

In case of categorical features, however, we introduce two masking mechanisms.
Firstly, we use $\bs{\mu}$ to denote the missingness of $\bs{\xi}$, i.e.,
\begin{align}
    \mu_j = \begin{cases}
        0 ~~\text{if}~ \xi_j ~\text{is missing}, \\
        1 ~~\text{otherwise}
    \end{cases} \label{eq:def_mu}
\end{align}

Once the features are binary and fuzzy coded, we construct another mask vector as:
\begin{align}
    \bs{m}_j = \begin{cases}
        \bs{0} ~~\text{if}~ \xi_j ~\text{is missing}, \\
        \bs{1} ~~\text{otherwise}
    \end{cases} \in [0, 1]^{q_j}, \label{eq:def_mask}
\end{align}
where we denote a \emph{vector} of 0's and 1's using $\bs{0}$ and $\bs{1}$, respectively.
It can be interpreted as simply repeating $\mu_j$ for $q_j$ times for the $j$-th feature.
The rationale for these two kinds of masking is that the discriminator's prediction is in fact equivalent to the missingness of the data.
For real-valued features discussed in \cite{yoon2018gain}, one could simply reuse the masking vector as the target of the discriminator.
But for categorical features that are coded as binary or fuzzy binary, doing so would imply making a prediction for each single \emph{category} instead of each \emph{feature}.
In the following introduction to the generator and discriminator, we shall give a more detailed explanation.

Analogously to the construction of $\bs{\bar{x}}$, we have the concatenation of $\bs{m}_j$'s:
\begin{align}
    \bs{\bar{m}} = [\bs{m}_1, \bs{m}_2, ..., \bs{m}_p\mathcal{]} =  [\bs{m}_j]_{j=1}^{p} \in [0, 1]^{\sum_{j=1}^p q_j}
\end{align}

\paragraph{The generator} 
The generator takes as input i) the fuzzy binary coded feature vector $\bs{\bar{x}}$ that is expected to contain missing values, ii) the equally sized mask vector $\bs{\bar{m}}$ and iii) a random vector $\bar{\bs{r}} = [\bs{r}_j]_{j=1}^{p}$ functioning as seeds. The generator produces as output a single vector denoted $\bs{\bar{g}}$ that is supposed to contain imputed missing values in $\bs{\bar{x}}$:
\begin{align}
    \bs{\bar{g}} = G(\bs{\bar{x}}, \bs{\bar{m}}, \bs{\bar{r}}).
\end{align}

%\textsc{comment: Is $g$ a full feature vector, predicting both the missing and the complete? What is the dimension of $g$? The figure makes this clear! } 

\begin{figure}[htbp]
\includegraphics[page=2 ,trim=7cm 0cm 7cm 0cm,clip=true,scale=0.45]{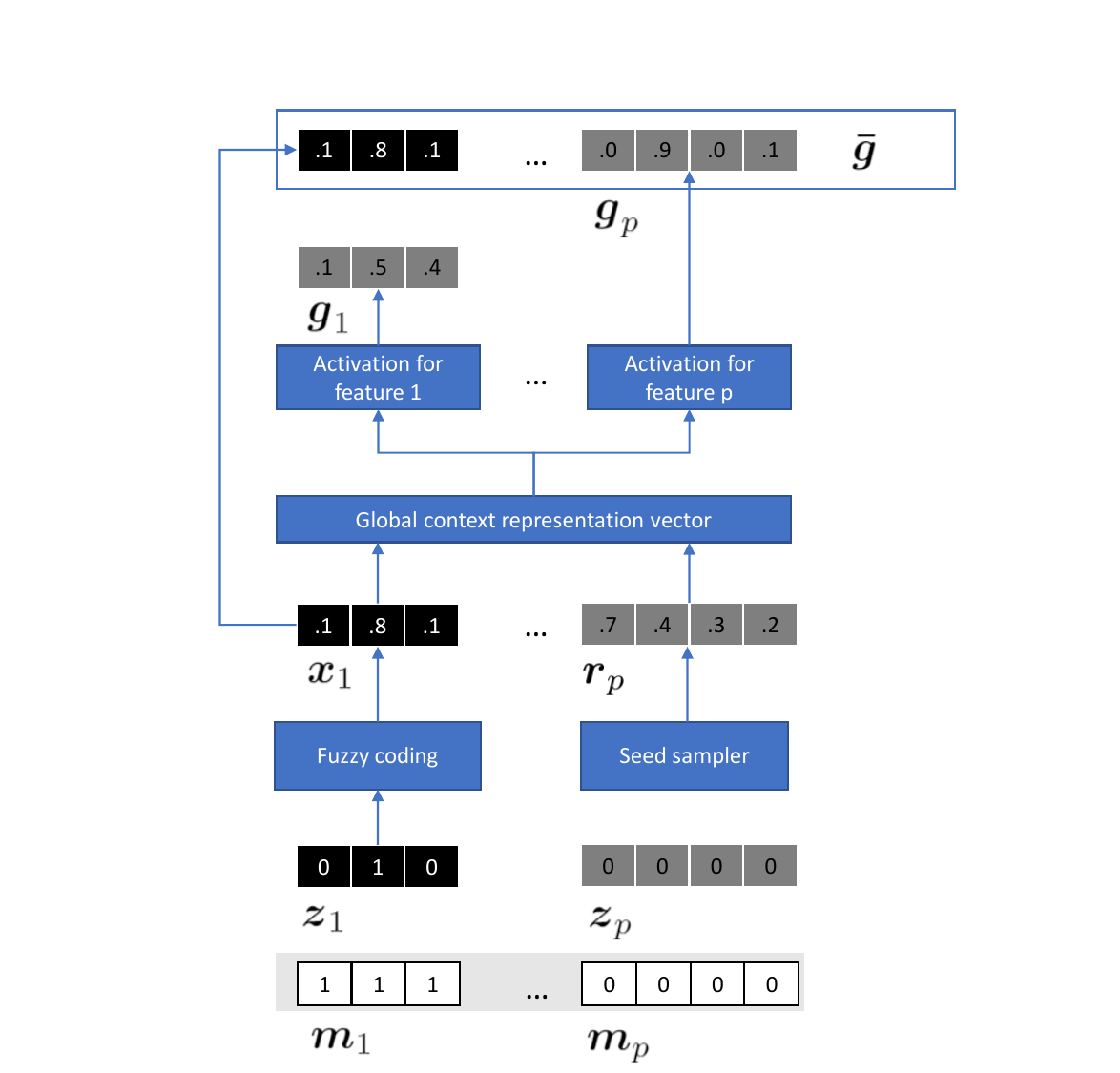}
\caption{A detailed illustration of the generator in \textit{categorical GAIN} architecture. %}  % \textsc{(is this the name of our new approach? Otherwise we should introduce one).} COmment: I would simply call our approach categorical GAIN, because our modification is relatively minor. But when it comes to this fuzzy coding trick, I would suggest writing another more methodology paper for an ML conference. What do you think??
As an example, we visualize two categorical features that are binary coded as $\bs{z}_1$ and $\bs{z}_p$.
The first is labeled as observed while the second as missing in $\bs{m}_1$ and $\bs{m}_p$.
Therefore, the observed binary input of $\bs{z}_1 = [0, 1, 0]$ is transformed into a fuzzy representation of $\bs{x}_1 = [0.1, 0.8, 0.1]$.
And the seed sampler fills these positions with random values in $\bs{r}_p$.
On the output side, the generated values for the first feature are replaced by the original, fuzzy codes, since the true values are observed, i.e., $\bs{g}_1 := \bs{x}_1$.
Only the generated values for the other feature $\bs{g}_p$ are exposed to the discriminator.
The concatenation of the overall generator output is denoted as $\bar{\bs{g}}$.}
\label{fig:architecture_G}
\end{figure}

Specifically in our implementation, we build as generator a neural network with 3 hidden layers:
\begin{align}
    \bs{h}^G_1 &= \text{relu}(\bs{W}^G_1 \cdot [\bs{\bar{x}} + (1-\bs{\bar{m}})\circ\bs{\bar{r}}, ~\bs{\bar{m}}] + \bs{b}^G_1) \label{eq:1-1}\\
    \bs{h}^G_2 &= \text{relu} (\bs{W}^G_2 \cdot \bs{h}^G_1 + \bs{b}^G_2) \\
    \bs{h}^G_3 &= \text{relu} (\bs{W}^G_3 \cdot \bs{h}^G_2 + \bs{b}^G_3) \\
    \bs{g}_j &= \sigma(\bs{W}^G_o(j) \cdot \bs{h}^G_3 + \bs{b}^G_o(j)) ~\forall j \in [1, p] \\
    \bs{\bar{g}} &= \bs{\bar{m}} \circ \bs{\bar{x}} + (1-\bs{\bar{m}}) \circ [\bs{g}_j]_{j=1}^{p} \label{eq:1-2}
\end{align}
As proposed by \cite{yoon2018gain}, the operation carried out in Eq. \eqref{eq:1-1} first fills the missing values in $\bs{\bar{x}}$ with random seeds $\bs{\bar{r}}$, before feeding it to the neural network.
The hidden layers $\bs{h}_1^G, \bs{h}_2^G, \bs{h}_3^G$ extract hierarchically the global context information from the input.
In the last layer, we define for each categorical feature $j$ a specific classification model.
Depending on the distribution assumption of the feature, the activation function can be either sigmoid or softmax, both of which are denoted using $\sigma$ for the sake of simplicity.
In Eq. \eqref{eq:1-2}, the outputs from all $p$ activation functions are concatenated as $[\bs{g}_j]_{j=1}^{p}$.
And if a specific feature is in fact not missing, the generated values are replaced by the real values.
Similar to the Multiple Imputation by Chained Equations \cite{stuart2009multiple}, this generator in fact attempts to approximate the real distribution $\bs{\pi}_{j^*}$ of each missing variable $\bs{X}_{j^*}$ conditioned on all other observed features $\bs{x}_j$, i.e.,
\begin{align}
    \hat{\bs{\pi}}_{j^*} = \bs{g}_{j^*} = \mathbb{P}(\bs{m}_{j^*}=\bs{1} ~|~ \{\bs{X}_{j} = \bs{x}_j\}_{\forall j: \mu_j = 1})
\end{align}
This architecture is illustrated in Fig. \ref{fig:architecture_G} with only two categorical features as examples.

%\textsc{Comment: you need to mention in the main text that you call your approach \textit{categorical GAIN}}
%\textsc{Comment: Can you visually indicate that $g_p$ is a NN or something?}

\paragraph{The discriminator}
Like the generator, the discriminator also consumes two input vectors.
The first input is the concatenated output of the generator $\bs{\bar{g}}$.
The second input is a hint vector as proposed by \cite{yoon2018gain}, which can be interpreted as a \emph{masked mask vector}:
For each data instance, one randomly samples a predefined portion of features and sets the corresponding entries in the mask vector $\bs{\bar{m}}$ to be 0.5.
On the output side of the discriminator, we have again an concatenated vector $\hat{\bs{\mu}} = [\hat{\mu}]_{j=1}^{p}$.
Each $\hat{\mu}_j$ is a point estimate of $\mu_j$ as defined in Eq. \eqref{eq:def_mu}, indicating whether the $j$-th feature in the input, denoted as $\bs{g}_j$ is generated or real,
\begin{align}
    \hat{\bs{\mu}}_{j^*} = \mathbb{P}(\bs{g}_{j^*} \text{is real} ~|~ \{\bs{g}_j\}_{\forall j: j \neq j^*}).
\end{align}
Generally, we can describe the discriminator as
\begin{align}
    \hat{\bs{\mu}} = D(\bs{\bar{g}}, \bs{\bar{h}}).
\end{align}

Specifically for our experiments, we have a neural network with two hidden layers:
\begin{align}
    \bs{h}^D_1 &= \text{relu}(\bs{W}^D_1 \cdot [\bs{\bar{g}}, \bs{\bar{h}}] + \bs{b}^G_1) \\
    \bs{h}^D_2 &= \text{relu} (\bs{W}^D_2 \cdot \bs{h}^D_1 + \bs{b}^D_2) \\
    \hat{\mu}_j &= \sigma(\bs{w}^D_o(j)^T \cdot \bs{h}^G_3 + b^D_o(j)) ~\forall j \in [1, p]
\end{align}
In parallel to the architecture of the generator,  the first two hidden layers represent the global context information, while the last layer contains $p$ logistic regression models.
Each of them attempts to predict whether the $j$-th feature in the input $\bs{g}_j$ is generated.
In the original GANs, each input vector to the discriminator is typically either generated or real.
In GAIN, however, one input vector to the discriminator may contain generated and real data simultaneously, and the discriminator performs multiple predictions correspondingly.

%\textsc{(comment: this is very detailed. Maybe the structure should indicate that the nontechnical reader can skip this: in the appendix, this is probably too drastic ... but maybe there is another solution)}
% Comment: you mean the definition of the NN? How would you formulate that "the nontechnical reader should skip"?? Any suggestion would be appreciated :)

In the original setting in \cite{yoon2018gain}, where the features are of real values, the training target of the discriminator is in fact identical with the mask vector.
In case of categorical features, however, one should not directly utilize the mask vector $\bs{\bar{m}}$ as training target.
Because in order to mask a (fuzzy) binary coded vector $\bs{x}_j$ completely, we have to define a same sized vector $\bs{m}_j$.
Training a discriminator that attempts to recover every element in $\bs{m}_j$ is in fact a prediction for each \emph{category} instead of \emph{feature}.
To this end, we propose to train the discriminator so that each $\hat{\mu}_j$ would approximate $\mu_j$ as in Eq. \eqref{eq:def_mu} for all real data.
The generator, on the other hand, should make the discriminator assign a $\hat{\mu}_j$ that is close to $1-\mu_j$ to all generated values.

The hint mechanism is also a crucial component in training the discriminator.
Once a subset of entries in the mask vector is set to a neutral value of 0.5, the discriminator is enforced to predict whether the corresponding values in $\bs{\bar{g}}$ are real or generated.
Such prediction is supposed to rely on other entries in $\bs{\bar{g}}$ that are provided to discriminator.
The proportion of features that are neutralized in the hint vector therefore controls the amount of information from which the discriminator is supposed to learn the decision.
In order to see that one could consider two extreme cases:
With the proportion close to 1, the discriminator would attempt to perform prediction for a large amount of features in $\bs{\bar{g}}$, based on very few features that are denoted as either real or generated.
This could be a challenging task for the discriminator and, more importantly, the discriminator may not learn to build the prediction based on the dependency between features.
With a proportion that is close to 0, the hint vector becomes almost identical to the mask vector.
In the original setting in \cite{yoon2018gain}, where features are of real values and the mask vector is in fact the prediction target of the discriminator, having two almost identical vectors as input and output of a neural network would cause the discriminator to simply learn an identity function, not being able to tell the difference between real and generated data.
This is slightly less of a problem in case of categorical features, because as stated above, our mask vector as input to the discriminator and training target are not exactly identical, although they contain the same information on the missingness of the data.
To this end, for experiments, we include the hint mechanism and use a relatively small proportion of 0.1.
This reveals $90\%$ of available information of the data missingness to the discriminator, which is encouraged to build its prediction based on the dependency among features.

\begin{figure}[ht]
\includegraphics[page=3,trim=7cm 4cm 7cm 1cm,clip=true,scale=0.45]{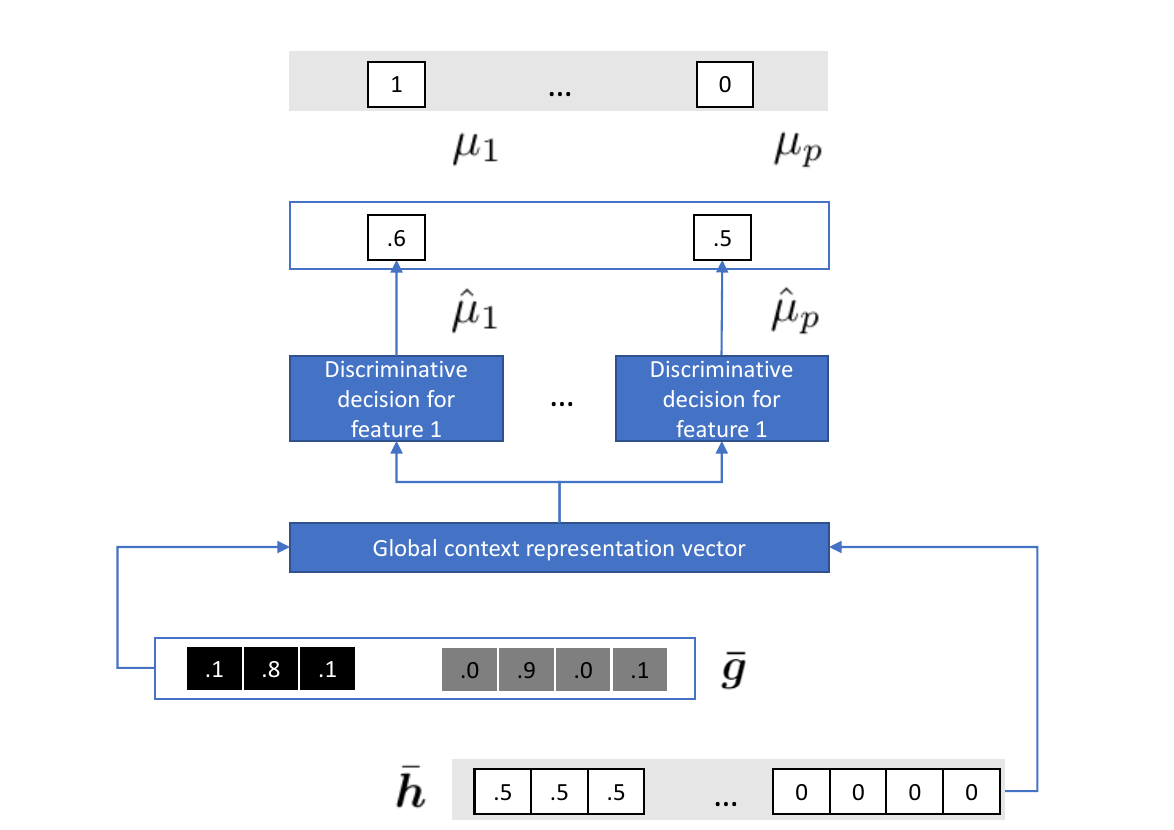}
\caption{A detailed illustration of the discriminator in categorical GAIN architecture. The input to the discriminator is the concatenated output $\bs{\bar{g}}$ from the generator, as well as the hint vector $\bs{\bar{h}}$.
The output of the discriminator here consists of 2 scalars of $\hat{\mu}_1$ and $\hat{\mu}_p$. They are trained against the scalars $\mu_1$ and $\mu_p$, encoding the missingness of both features respectively.}
\label{fig:architecture_D}
\end{figure}

The hint vector also has to be adjusted for categorical features.
Similar to the mask vectors, we define for each categorical feature $j$ a hint vector $\bs{h}_j$ that consists of exclusively either 0 or 1, and denote the concatenation of all hint vectors as $\bs{\bar{h}} = [\bs{h}_j]_{j=1}^{p}$.
The proposed approach in \cite{yoon2018gain} would imply masking the missingness information for each \emph{category} instead of each \emph{feature}.
To this end, we propose to first sample a subset out of the $p$ features, and set the entire corresponding hint vectors to be 0.5, i.e. $\bs{h}_j = \bs{0.5}$, as can be seen in the illustration in Fig. \ref{fig:architecture_D}

\paragraph{Loss functions}
Similar to the original GANs framework as in Eq. \eqref{eq:preliminary_gan}, GAIN also contains two adversarial loss functions $loss_D$ and $loss_G$:
\begin{align}
    \textit{loss}_D &= -\Sigma_j \left( \mu_j \cdot \log(\hat{\mu}_j) + (1-\mu_j) \cdot \log(1-\hat{\mu}_j) \right) \label{eq:2_1} \\
    \textit{loss}_G &= -\Sigma_j (1-\mu_j) \cdot \log(\hat{\mu}_j) \label{eq:3_1}
\end{align}
%\textsc{In the equations: use \textit{loss}}
% Comment: why? why not differentiating between D and G??
The discriminator adjusts itself to make correct classification by minimizing the $loss_D$ in Eq. \eqref{eq:2_1}.
This objective forces the discriminator to produce large $\hat{\mu}_j$ if $\mu_j=1$, indicating that $\bs{x}_j$ is real.
The generator learns to fool the discriminator by minimizing $loss_G$ in Eq. \eqref{eq:3_1}.
This loss is adversarial to the second additive term in the discriminator loss.
The generator encourages the discriminator to assign large probability $\hat{\mu}_j$ to features where $\mu_j = 0$, implying that the generated data should be classified as real.

As defined in Eq. \eqref{eq:1-2}, once a feature $j$ is observed instead of missing, whatever is generated by the generator gets replaced by the actually observed values.
The weight parameters responsible for these features will not get gradient signals for this specific training sample.
Therefore, \cite{yoon2018gain} proposes a new loss function that measures the similarity between generated and the observed feature values.
In case of real-valued features this loss could be realized as mean-squared error.
In our case, we apply the log-loss to measure the distance between probabilities and binary codes:
\begin{align}
	loss_{sim} &= \Sigma_j \bs{m}_j^T (-\bs{x}_j)\log(\bs{g}_j) \label{eq:3_2}.
\end{align}

This loss mechanism implies that, in case a feature is observed, the generator should learn to reproduce it based on all other observed features; and in case a feature is missing, the adversarial training forces the generator to produce values that the discriminator would believe to be real.

\iffalse
\begin{figure}[ht]
\centerline{
\includegraphics[page=1,trim=5cm 3cm 10cm 1cm,clip=true,scale=0.45]{graphics.pdf}}
\caption{An overview of the complete GAIN architecture\cite{yoon2018gain}}
\label{fig:architecture_overview}
\end{figure}
\fi

In comparison to the original GAIN architecture in \cite{yoon2018gain}, there are three adjustments that we propose for categorical features.
First, the output activation function in the generator: in order to take into account the discrete distribution of the data features, we apply softmax or sigmoid activation functions instead of linear activation.
Second, the target variable of the discriminator: In case of real valued features, the discriminator only needs to predict the mask vector $\bs{\bar{m}}$ which has the same shape as the feature vector $\bs{\bar{x}}$. This is because each element in the mask vector can represent the missingness of the corresponding feature. However, in order to encode the missingness of a feature containing multiple categories $\bs{x}_j$, it is unnecessary for the discriminator to recover the corresponding mask vector $\bs{m}_j$, since all values in this vector are either all 0's or all 1's. Instead, it is much more efficient to train the discriminator to predict the scalar $\mu_j$.
Thirdly, due to the same reason, the hint mechanism also has to be defined on the level of feature instead of categories.
In other words, for a feature $j$ we initialize a vector $\bs{h}_j$ from $\bs{m}_j$, and set all elements to be $0.5$ if necessary.

%For each categorical feature $x_j$, the generator models its probabilistic distribution dependent on all other features:
%\begin{align}
%    \mathbb{P}(x_j = c_j | \{x_{j'}\}_{\forall j' \neq j} )
%\end{align}

\section{Experiments} \label{sec:experiments}
In this section, we provide experiments conducted on two datasets.
The first dataset is publicly available and a well known benchmark for breast cancer classification based on categorical features.
The second dataset is provided by the PRAEGNANT study \cite{fasching2015biomarkers}, a Germany-wide clinical study for breast cancer research.

Please recall that we perform fuzzy binary coding of the categorical features and our generator produces values that range between 0 and 1.
We recover the binary codes applying $I(\bs{x}_j \geq 0.5)$ for multilabel and $I(\bs{x}_j = \max(\bs{x}_j))$ for multiclass features as a post processing step.
Because, as discussed in subsection \ref{subsec:fuzzy_coding}, the encoded categorical information is always retained after the fuzzy binary coding and can be recovered completely.

\subsection{Experiments on a public dataset}
The breast cancer dataset is available on UCI data repository \cite{Dua:2017}.
It contains 9 multiclass features (Tab. \ref{tab:features_public}) observed on 286 patient cases.
The prediction target is to differentiate between 201 recurrence and 85 no-recurrence cases of the cancer.
\begin{table}[ht]
    \centering
    \begin{tabular}{|l|l|} \hline
         Feature & \#Categories \\ \hline \hline
         age & 6 \\
         menopause & 3 \\
         tumor-size & 11\\
         inv-nodes & 7 \\
         node-caps & 2 \\
         deg-malig & 3 \\
         breast & 2 \\
         breast-quad & 5 \\
         irradiat & 2 \\ \hline
    \end{tabular}
    \vspace{0.25cm}
    \caption{Patient features from the UCI breast cancer dataset }
    \label{tab:features_public}
\end{table}

We perform 5-fold cross-validation on the complete dataset, applying logistic regressions with ridge regularization (Tab. \ref{tab:check_public}) and report the prediction accuracy and AUROC scores.
As sanity check we also provide these scores produced by random and most popular predictions, the latter of which constantly produces the frequency of the label class in the training set.
\begin{table}[ht]
    \centering
    \begin{tabular}{| r | l | l | } \hline
    Methods & Accuracy & AUROC \\ \hline \hline
    Random prediction                & $0.516 \pm 0.051$ & $0.484 \pm 0.053$ \\
    Most popular prediction          & $0.707 \pm 0.051$ & $0.500 \pm 0$  \\
    Prediction on complete data      & $0.737 \pm 0.056$ & $0.721 \pm 0.051$ \\ \hline
    \end{tabular}
    \vspace{0.2cm}
    \caption{Sanity checks for the prediction task on the UCI breast cancer dataset}
    \label{tab:check_public}
\end{table}

For each cross-validation split, we randomly mask 10\%, 20\%, 30\% 40\% and 50\% of the features.
We then apply different imputation approaches to recover the masked values.
Note that the imputation model is only trained on the training set, and applied on both training and test sets, in order to simulate a realistic setting.
The predictive model is then trained on \emph{imputed} training set and validated on the \emph{imputed} test set.

As the first baseline method we implement a low-rank reconstruction model using SVD.
Assuming $\bs{X}_{tr}$ and $\bs{X}_{te}$ as training and test sets containing missing values, we compose the former as $\bs{X}_{tr} = \bs{U} \bs{D} \bs{V}^T$, and impute the training and test sets as $\Tilde{\bs{X}}_{tr} = \bs{U}_r \bs{D}_r \bs{V}_r^T$ and $\Tilde{\bs{X}}_{te} = \bs{X}_{te} (\bs{D}_r \bs{V}_r^T)^{\dagger}(\bs{D}_r \bs{V}_r^T)$, respectively.
Here we denote the the low rank representation of $\bs{U}, \bs{D}$ and $\bs{V}^T$ using $\bs{U}_r, \bs{D}_r$ and $\bs{V}_r^T$ with a specific rank $r$.
Please note that we do not perform any \textit{argmax} to the reconstructed values. 

%\textsc{comment: I am not sure I understand how you use SVD here: Do you first impute all missings as zero and see what happens in the reconstruction?  Do you then do a max in the multiclass setting? How about in the multilabel?}

The same ranks also apply to the second baseline model, which is an auto-encoder with non-linear $\tanh$ activation for the hidden layer.
We summarize all prediction performances in term of accuracy and AUROC scores in Tab. \ref{tab:res_public} as average and standard deviation of the 5 cross-validation splits.
For both baseline models we conduct experiments using 4 different ranks, i.e., the size of the hidden layer in AE, of 4, 8, 16 and 32, and report the best results.
For the categorical GAIN model we perform 100-fold multiple imputation.

\begin{table}[ht]
    \centering
    \begin{tabular}{| r | r | l | l | } \hline
    & Methods & Accuracy & AUROC \\ \hline \hline
    \multirow{4}{*}{10\%} &
No imputation & $0.718 \pm 0.067$ &$0.66 \pm 0.11$ \\
&Avg imputation & $0.744 \pm 0.05$ &$0.639 \pm 0.088$ \\
&SVD reconstruction & $\textbf{0.776} \pm 0.062$ &$0.689 \pm 0.114$ \\
&Auto-encoder & $0.751 \pm 0.047$ &$0.652 \pm 0.089$ \\
&Categorical GAIN & $0.739 \pm 0.066$ &$\textbf{0.697} \pm 0.098$ \\
\hline \hline
    \multirow{4}{*}{20\%} &
No imputation & $0.711 \pm 0.039$ &$0.634 \pm 0.07$ \\
&Avg imputation & $0.707 \pm 0.036$ &$0.671 \pm 0.065$ \\
&SVD reconstruction & $\textbf{0.747} \pm 0.046$ &$0.664 \pm 0.082$ \\
&Auto-encoder & $0.729 \pm 0.051$ &$0.636 \pm 0.038$ \\
&Categorical GAIN & $0.71 \pm 0.046$ &$\textbf{0.697} \pm 0.087$ \\
    \hline \hline
    \multirow{4}{*}{30\%} &
No imputation & $0.726 \pm 0.031$ &$0.644 \pm 0.086$ \\
&Avg imputation & $0.729 \pm 0.04$ &$0.665 \pm 0.071$ \\
&SVD reconstruction & $0.726 \pm 0.053$ &$0.689 \pm 0.083$ \\
&Auto-encoder & $0.726 \pm 0.038$ &$0.641 \pm 0.053$ \\
&Categorical GAIN & $\textbf{0.737} \pm 0.032$ &$\textbf{0.704} \pm 0.042$ \\
\hline \hline
    \multirow{4}{*}{40\%} &
No imputation & $0.678 \pm 0.052$ &$0.686 \pm 0.058$ \\
&Avg imputation & $0.708 \pm 0.027$ &$0.54 \pm 0.066$ \\
&SVD reconstruction & $\textbf{0.751} \pm 0.026$ &$\textbf{0.709} \pm 0.059$ \\
&Auto-encoder & $0.737 \pm 0.033$ &$0.638 \pm 0.054$ \\
&Categorical GAIN & $0.7 \pm 0.017$ &$0.686 \pm 0.051$ \\
    \hline \hline
    \multirow{4}{*}{50\%} &
No imputation & $0.701 \pm 0.044$ &$0.607 \pm 0.091$ \\
&Avg imputation & $0.704 \pm 0.044$ &$0.632 \pm 0.057$ \\
&SVD reconstruction & $\textbf{0.747} \pm 0.029$ &$0.665 \pm 0.063$ \\
&Auto-encoder & $0.74 \pm 0.034$ &$0.635 \pm 0.041$ \\
&Categorical GAIN & $0.713 \pm 0.025$ &$\textbf{0.72} \pm 0.044$ \\
    \hline
    \end{tabular}
    \vspace{1cm}
    \caption{Prediction performances on imputed UCI breast cancer dataset using different approaches}
    \label{tab:res_public}
\end{table}

In term of accuracy, SVD reconstruction turns out to be more effective for this dataset, achieving the best accuracy in 4 out of 5 settings of masking proportions.
In term of AUROC, categorical GAIN achieves 4 out of 5 cases.
It is therefore interesting to note that for this dataset, the SVD decomposition does not take into account the the fact that the feature values are in fact binary
And yet the SVD reconstruction achieves comparable performances as categorical GAIN.
This relatively simple technique, as well as many approaches that it has inspired, are widely applied in recommender systems and knowledge graph, where the most essential task is the completion of matrices and tensors.
Therefore, it could very well present a simple and effective solution for data imputation as well.
However, one should also note that the label distribution in this dataset is relatively unbalanced (201:85).
Consequently, the most popular prediction as in Tab. \ref{tab:check_public} can already reach $70\%$ accuracy.
And even with complete data the prediction model cannot improve beyond $73.7\%$.
The AUROC, in contrast, seems to be a more informative and convincing measurement, because the prediction on complete data achieves $72\%$ while the most popular prediction $50\%$.
Therefore, for this dataset, ROC seems to be a more reliable means to measure the prediction quality.

\subsection{Experiments on the PRAEGNANT dataset}
\subsubsection{Cohort and Features}
For our experiment, we extract EHR data on 1234 patients with metastatic breast cancer who have met the first line of treatment from the PRAEGNANT study network \cite{fasching2015biomarkers}.
We build our predictive models based on features that are clinically relevant, as well as those that are based on an earlier study \cite{yang2018explaining} aiming at automatically inferring the feature relevance in EHR data.
The features included are listed in Tab. \ref{tab:features_private}.
We have 10 multiclass features and 9 multilabel featues, both of which are fuzzy-binary coded.
The one numeric feature is normalized between 0 and 1. 
Features such as current metastasis, metastasis estrogen receptor, metastasis progesterone receptor, AE/SAE and ECOG life status were originally temporal features.
We aggregate and normalize these w.r.t the time dimension as in \cite{esteban2015predicting}.
For these patients it is especially important for the physicians to decide, whether they should receive antihormone therapy or chemo therapy.
The recorded clinical decision serve as ground truth, i.e. the target of our prediction.
750 of the 1234 patients have received antihormone, and the rest chemo therapy.

\begin{table}[ht]
    \centering
    \begin{tabular}{| l | l | } \hline
         Multiclass features                & \#Categories \\ \hline
         Staging at breast                  & 15     \\  % pt
         Staging at axilla                  & 8      \\  % n
         Ever received antihormone therapy  & 8      \\  % antihormonjn
         Ever received chemo therapy        & 8      \\  % chemojn
         Metastasis by diagnostics          & 5      \\  % metprimary
         Tumor estrogen receptor status     & 4      \\  % sub_patbi_er
         Tumor progesterone receptor        & 4      \\  % sub_patbi_pgr
         Immunohistochemistry for HER2      & 6      \\  % her2ihc
         Tumor grading                      & 5      \\  % grading1
         KI67                               & 3      \\  % ki67
         \hline \hline
         Multilabel features                & \#Categories \\ \hline
         Staging of metastasis              & 10     \\  % metprimaryort
         Location of earlier metastasis     & 14     \\  % metlunge, ..., other
         Current metastasis                 & 4      \\  % metdoku
         Metastasis estrogen receptor       & 3      \\  % ermetlok
         Metastasis progesterone receptor   & 3      \\  % prmetlok
         HER2 IHC                           & 5      \\  % her2ihcmetlok
         Metastasis grading                 & 4      \\  % grademetlok
         AE/SAE                             & 20     \\
         ECOG life status                   & 4      \\
         \hline \hline
         Numerical features                & \#Dimension \\ \hline
         Age                                & 1      \\  % age
         \hline
    \end{tabular}
    \vspace{0.25cm}
    \caption{Patient features from the PRAEGNANT study. }
    \label{tab:features_private}
\end{table}

Here we apply almost exactly the same experimental setting as with the public dataset, except that, considering the feature space of higher dimension, we train the SVD and auto-encoder imputation models with an additional rank of 64.

\subsubsection{Experimental Results}
\begin{table}[ht]
    \centering
    \begin{tabular}{| r | l | l | } \hline
    Methods & Accuracy & AUROC \\ \hline \hline
    Random prediction                & $0.516 \pm 0.029$ & $0.526 \pm 0.041$ \\
    Most popular prediction          & $0.607 \pm 0.046$ & $0.500 \pm 0$     \\
    Prediction on complete data      & $0.710 \pm 0.029$ & $0.774 \pm 0.039$ \\ \hline
    \end{tabular}
    \vspace{0.2cm}
    \caption{Sanity checks for the prediction task on the PRAEGNANT dataset}
    \label{tab:check_private}
\end{table}

In Tab. \ref{tab:check_private} we could see there is a large improvement from most popular prediction to the prediction on complete data in term of both accuracy and AUROC.

\begin{table}[ht]
    \centering
    \begin{tabular}{| r | r | l | l | } \hline
    & Methods & Accuracy & AUROC \\ \hline \hline
    \multirow{4}{*}{10\%} &
No imputation & $0.674 \pm 0.017$ &$0.718 \pm 0.016$ \\
&Avg imputation & $0.689 \pm 0.008$ &$\textbf{0.727} \pm 0.024$ \\
&SVD reconstruction & $\textbf{0.7} \pm 0.015$ &$0.645 \pm 0.011$ \\
&Auto-encoder & $0.609 \pm 0.023$ &$0.506 \pm 0.022$ \\
&Categorical GAIN & $0.645 \pm 0.012$ &$0.725 \pm 0.024$ \\
    \hline \hline
    \multirow{4}{*}{20\%} &
No imputation & $0.669 \pm 0.014$ &$0.69 \pm 0.011$ \\
&Avg imputation & $\textbf{0.684} \pm 0.015$ &$0.707 \pm 0.014$ \\
&SVD reconstruction & $0.663 \pm 0.025$ &$0.621 \pm 0.032$ \\
&Auto-encoder & $0.609 \pm 0.021$ &$0.496 \pm 0.03$ \\
&Categorical GAIN & $0.649 \pm 0.016$ &$\textbf{0.716} \pm 0.022$ \\
    \hline \hline
    \multirow{4}{*}{30\%} &
No imputation & $0.645 \pm 0.03$ &$0.696 \pm 0.018$ \\
&Avg imputation & $0.658 \pm 0.039$ &$0.695 \pm 0.021$ \\
&SVD reconstruction & $0.662 \pm 0.04$ &$0.599 \pm 0.018$ \\
&Auto-encoder & $0.609 \pm 0.043$ &$0.528 \pm 0.017$ \\
&Categorical GAIN & $\textbf{0.665} \pm 0.018$ &$\textbf{0.723} \pm 0.01$ \\
\hline \hline
    \multirow{4}{*}{40\%} &
No imputation & $0.652 \pm 0.008$ &$0.663 \pm 0.009$ \\
&Avg imputation & $0.643 \pm 0.012$ &$0.66 \pm 0.014$ \\
&SVD reconstruction & $0.658 \pm 0.01$ &$0.6 \pm 0.017$ \\
&Auto-encoder & $0.608 \pm 0.017$ &$0.494 \pm 0.034$ \\
&Categorical GAIN & $\textbf{0.666} \pm 0.017$ &$\textbf{0.711} \pm 0.015$ \\
    \hline \hline
    \multirow{4}{*}{50\%} &
No imputation & $0.635 \pm 0.027$ &$0.646 \pm 0.029$ \\
&Avg imputation & $0.649 \pm 0.041$ &$0.643 \pm 0.038$ \\
&SVD reconstruction & $0.644 \pm 0.015$ &$0.566 \pm 0.018$ \\
&Auto-encoder & $0.608 \pm 0.013$ &$0.509 \pm 0.038$ \\
&Categorical GAIN & $\textbf{0.654} \pm 0.05$ &$\textbf{0.705} \pm 0.029$ \\
    \hline
    \end{tabular}
    \vspace{1cm}
    \caption{Prediction performances on imputed PRAEGNANT dataset using different approaches}
    \label{tab:res_private}
\end{table}

In Tab. \ref{tab:res_private} we could see that, the advantage of categorical GAIN only becomes visible as the masking proportion increases.
For smaller proportion like $10\%$ and $20\%$, simpler methods such as average imputation and SVD shows superior performances.
With a proportion larger than $30\%$, categorical GAIN outperforms all other methods and the improvement grows with masking proportion.
In term of AUROC, for instance, categorical GAIN can always achieve a score above $70\%$, while the other performance of other methods drop much faster as the proportion of missing data increases.
This agrees with findings in \cite{yoon2018gain}, that it is especially advantageous to apply GAIN to impute data in case of a relatively higher missing rate.

One might also hypothesize that the GAIN framework, consisting of relatively complex neural networks, profit from increasing number of training samples.
For a smaller dataset such as the public breast cancer dataset, it seems more reasonable to first experiment with simpler methods such as SVD reconstruction.
The GAIN approach, on the other hand, turns out to be more appropriate in case of large number of training samples and more complex feature dependencies.

\begin{figure}[htbp]\label{tab:losses_private}
\includegraphics[trim=0cm 0cm 0cm 0cm,clip=true,scale=0.25]{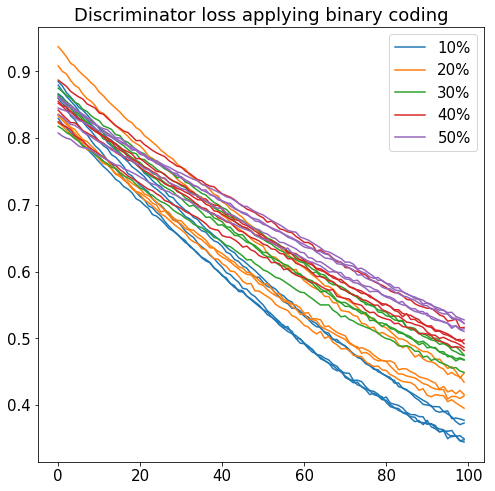}
\includegraphics[trim=0cm 0cm 0cm 0cm,clip=true,scale=0.25]{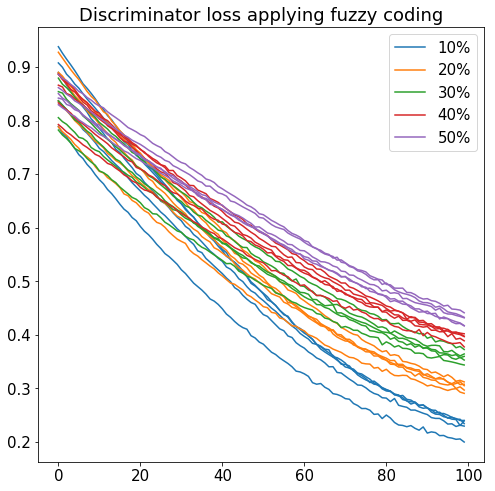} \\
\includegraphics[trim=0cm 0cm 0cm 0cm,clip=true,scale=0.25]{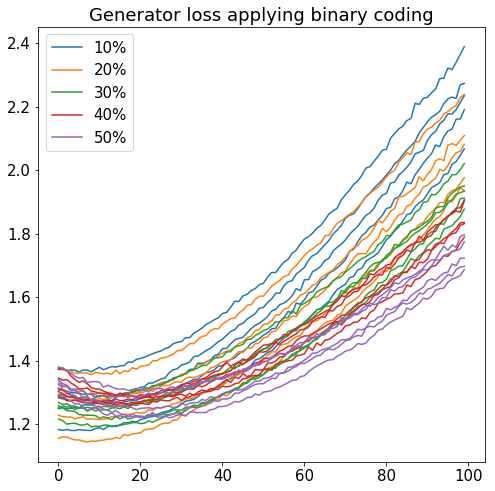}
\includegraphics[trim=0cm 0cm 0cm 0cm,clip=true,scale=0.25]{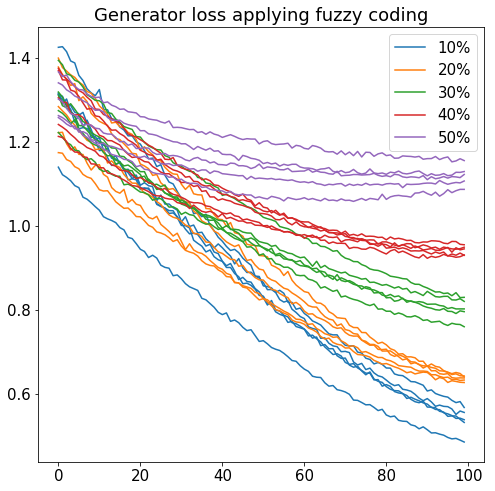}
\caption{Losses in adversarial training on the PREAGNANT dataset. X-axis: training epochs; Y-axis: adversarial loss. Above: Discriminator losses with binary coding (left) and fuzzy binary coding (right). Bottom: Generator losses with binary coding (left) and fuzzy binary coding (right).}
\label{fig:GLoss}
\end{figure}

We also present in Fig. \ref{fig:GLoss} the development of the losses of discriminator (top) and generator (bottom), trained on binary (left) and fuzzy coded (right) features.
In case of plain binary coded features, it is clear that the adversarial training fails since the generator loss increases, while the discriminator loss decreases constantly.
This implies that the discriminator can always tell the real data from generated ones.
Consequently, the generator cannot improve itself by learning to generate important characteristics in the feature distribution.
When we apply the fuzzy binary coding, in contrast, the generator can improve itself by lowering its loss, i.e., it gets harder and harder for the discriminator to make the decision.
In addition, as expected, varying proportion of missing data (masking) has impact on the adversarial training losses.
With larger proportion of missing data, the imputation task becomes more challenging and both discriminator loss and generator loss are expected to increase with larger proportion.
This verifies empirically our hypothesis, that, if one applies softmax as the final activation in the generator to generate categorical data, the adversarial training fails as the discriminator can learn to exploit the huge difference in the generated and real data.
This typically results in divergence of the adversarial training.
By re-coding the binary features in a fuzzy way while retaining the information, we enforce the real data to resemble what softmax would produce.
Thus we can make both discriminator and generator converge in training.

\section{Summary}
In this paper, we have proposed a Categorical Generative Adversarial Nets (\emph{Categorical GAIN}) for EHR data imputation, based on a framework that is originally designed for real values.
First, we have hypothesized that applying softmax functions as output activation in the generator directly often results in the discriminator exploiting the obvious difference between generated and real values. And the adversarial training typically ends up in divergence.
We have proposed to perform fuzzy coding of the binary values so that they resemble generated values while retaining the encoded information.
Secondly, we have performed multiple modifications in the architectures of both generator and discriminator, in order to handle the fuzzy binary coded features. 

We have compared our methods with a variety of benchmark methods on two EHR datasets.
We have simulated different proportions of missing data by masking out known values and then attempting to perform prediction tasks based on imputed data.
We could show that the more complex method of generative adversarial nets turned out to be advantageous in case of relatively higher missing rate and larger training data set, while the simpler methods such as SVD reconstruction and average imputation are more reliable to impute smaller proportion of missing data.

\section*{Acknowledgment}
The authors acknowledge support by the German Federal Ministry for Education and Research (BMBF),
funding project “MLWin” (grant 01IS18050).
%\vspace{3cm}
\begin{figure}[h]
\includegraphics[scale=0.08]{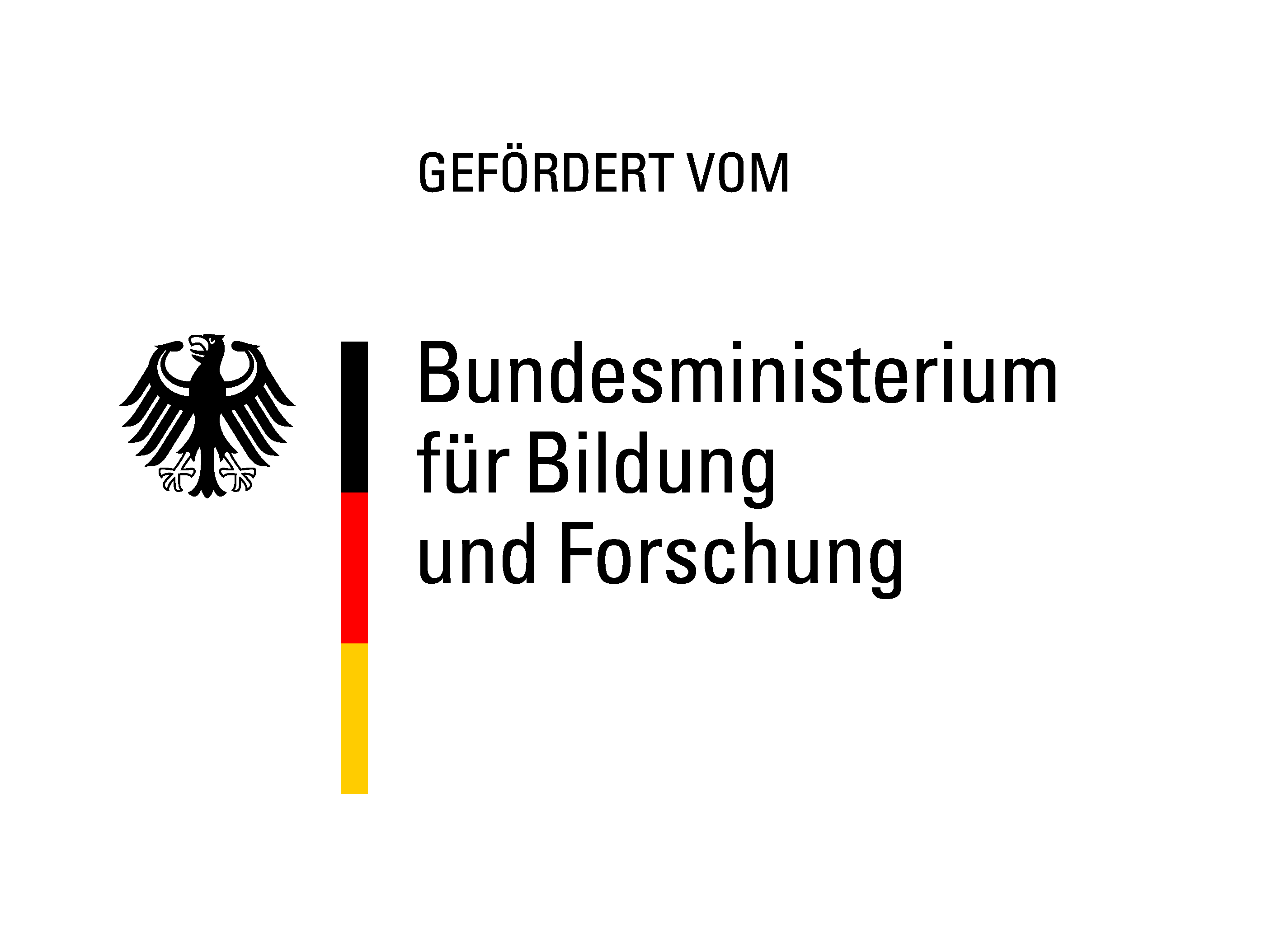}
\end{figure}

% stuff for the graphics:
%\begin{align}
%  \bs{\bar{m}} \\
%	\bar{\bs{h}}
%	\mu_p \\
%	\hat{\mu}_1 \\
%	\hat{\mu}_p \\
%	\bs{r}_p
%\end{align}

%\vspace{2cm}
% \bibliographystyle{IEEEtran}


\begin{thebibliography}{00}

\bibitem{arjovsky2017wasserstein}
Martin Arjovsky, Soumith Chintala, and L{\'e}on Bottou.
\newblock Wasserstein gan.
\newblock {\em arXiv preprint arXiv:1701.07875}, 2017.

\bibitem{choi2017generating}
Edward Choi, Siddharth Biswal, Bradley Malin, Jon Duke, Walter~F Stewart, and
  Jimeng Sun.
\newblock Generating multi-label discrete patient records using generative
  adversarial networks.
\newblock {\em arXiv preprint arXiv:1703.06490}, 2017.

\bibitem{Dua:2017}
Dua Dheeru and Efi Karra~Taniskidou.
\newblock {UCI} machine learning repository, 2017.

\bibitem{donders2006gentle}
A~Rogier~T Donders, Geert~JMG Van Der~Heijden, Theo Stijnen, and Karel~GM
  Moons.
\newblock A gentle introduction to imputation of missing values.
\newblock {\em Journal of clinical epidemiology}, 59(10):1087--1091, 2006.

\bibitem{esteban2017real}
Crist{\'o}bal Esteban, Stephanie~L Hyland, and Gunnar R{\"a}tsch.
\newblock Real-valued (medical) time series generation with recurrent
  conditional gans.
\newblock {\em arXiv preprint arXiv:1706.02633}, 2017.

\bibitem{esteban2015predicting}
Crist{\'o}bal Esteban, Danilo Schmidt, Denis Krompa{\ss}, and Volker Tresp.
\newblock Predicting sequences of clinical events by using a personalized
  temporal latent embedding model.
\newblock In {\em Healthcare Informatics (ICHI), 2015 International Conference
  on}, pages 130--139. IEEE, 2015.

\bibitem{fasching2015biomarkers}
PA~Fasching, SY~Brucker, TN~Fehm, F~Overkamp, W~Janni, M~Wallwiener, P~Hadji,
  E~Belleville, L~H{\"a}berle, F-A Taran, et~al.
\newblock Biomarkers in patients with metastatic breast cancer and the
  praegnant study network.
\newblock {\em Geburtshilfe und Frauenheilkunde}, 75(01):41--50, 2015.

\bibitem{goodfellow2014generative}
Ian Goodfellow, Jean Pouget-Abadie, Mehdi Mirza, Bing Xu, David Warde-Farley,
  Sherjil Ozair, Aaron Courville, and Yoshua Bengio.
\newblock Generative adversarial nets.
\newblock In {\em Advances in neural information processing systems}, pages
  2672--2680, 2014.

\bibitem{jang2016categorical}
Eric Jang, Shixiang Gu, and Ben Poole.
\newblock Categorical reparameterization with gumbel-softmax.
\newblock {\em arXiv preprint arXiv:1611.01144}, 2016.

\bibitem{kusner2016gans}
Matt~J Kusner and Jos{\'e}~Miguel Hern{\'a}ndez-Lobato.
\newblock Gans for sequences of discrete elements with the gumbel-softmax
  distribution.
\newblock {\em arXiv preprint arXiv:1611.04051}, 2016.

\bibitem{ledig2017photo}
Christian Ledig, Lucas Theis, Ferenc Husz{\'a}r, Jose Caballero, Andrew
  Cunningham, Alejandro Acosta, Andrew~P Aitken, Alykhan Tejani, Johannes Totz,
  Zehan Wang, et~al.
\newblock Photo-realistic single image super-resolution using a generative
  adversarial network.
\newblock In {\em CVPR}, volume~2, page~4, 2017.

\bibitem{li2017adversarial}
Jiwei Li, Will Monroe, Tianlin Shi, S{\'e}bastien Jean, Alan Ritter, and Dan
  Jurafsky.
\newblock Adversarial learning for neural dialogue generation.
\newblock {\em arXiv preprint arXiv:1701.06547}, 2017.

\bibitem{little2014statistical}
Roderick~JA Little and Donald~B Rubin.
\newblock {\em Statistical analysis with missing data}, volume 333.
\newblock John Wiley \& Sons, 2014.

\bibitem{mirkes2016handling}
Eugenij~Moiseevich Mirkes, Timothy~J Coats, Jeremy Levesley, and Alexander~N
  Gorban.
\newblock Handling missing data in large healthcare dataset: A case study of
  unknown trauma outcomes.
\newblock {\em Computers in biology and medicine}, 75:203--216, 2016.

\bibitem{mirza2014conditional}
Mehdi Mirza and Simon Osindero.
\newblock Conditional generative adversarial nets.
\newblock {\em arXiv preprint arXiv:1411.1784}, 2014.

\bibitem{raghunathan2001multivariate}
Trivellore~E Raghunathan, James~M Lepkowski, John Van~Hoewyk, and Peter
  Solenberger.
\newblock A multivariate technique for multiply imputing missing values using a
  sequence of regression models.
\newblock {\em Survey methodology}, 27(1):85--96, 2001.

\bibitem{rajeswar2017adversarial}
Sai Rajeswar, Sandeep Subramanian, Francis Dutil, Christopher Pal, and Aaron
  Courville.
\newblock Adversarial generation of natural language.
\newblock {\em arXiv preprint arXiv:1705.10929}, 2017.

\bibitem{rubin1996multiple}
Donald~B Rubin.
\newblock Multiple imputation after 18+ years.
\newblock {\em Journal of the American statistical Association},
  91(434):473--489, 1996.

\bibitem{rubin2004multiple}
Donald~B Rubin.
\newblock {\em Multiple imputation for nonresponse in surveys}, volume~81.
\newblock John Wiley \& Sons, 2004.

\bibitem{shrivastava2017learning}
Ashish Shrivastava, Tomas Pfister, Oncel Tuzel, Joshua Susskind, Wenda Wang,
  and Russell Webb.
\newblock Learning from simulated and unsupervised images through adversarial
  training.
\newblock In {\em CVPR}, volume~2, page~5, 2017.

\bibitem{stuart2009multiple}
Elizabeth~A Stuart, Melissa Azur, Constantine Frangakis, and Philip Leaf.
\newblock Multiple imputation with large data sets: a case study of the
  children's mental health initiative.
\newblock {\em American journal of epidemiology}, 169(9):1133--1139, 2009.

\bibitem{tresp1995efficient}
Volker Tresp, Ralph Neuneier, and Subutai Ahmad.
\newblock Efficient methods for dealing with missing data in supervised
  learning.
\newblock In {\em Advances in neural information processing systems}, pages
  689--696, 1995.

\bibitem{wells2013strategies}
Brian~J Wells, Kevin~M Chagin, Amy~S Nowacki, and Michael~W Kattan.
\newblock Strategies for handling missing data in electronic health record
  derived data.
\newblock {\em eGEMs}, 1(3), 2013.

\bibitem{williams1992simple}
Ronald~J Williams.
\newblock Simple statistical gradient-following algorithms for connectionist
  reinforcement learning.
\newblock {\em Machine learning}, 8(3-4):229--256, 1992.

\bibitem{yang2018explaining}
Yinchong Yang, Volker Tresp, Marius Wunderle, and Peter~A Fasching.
\newblock Explaining therapy predictions with layer-wise relevance propagation
  in neural networks.
\newblock In {\em 2018 IEEE International Conference on Healthcare Informatics
  (ICHI)}, pages 152--162. IEEE, 2018.

\bibitem{yoon2018gain}
Jinsung Yoon, James Jordon, and Mihaela van~der Schaar.
\newblock Gain: Missing data imputation using generative adversarial nets.
\newblock {\em arXiv preprint arXiv:1806.02920}, 2018.

\bibitem{yu2017seqgan}
Lantao Yu, Weinan Zhang, Jun Wang, and Yong Yu.
\newblock Seqgan: Sequence generative adversarial nets with policy gradient.
\newblock In {\em AAAI}, pages 2852--2858, 2017.

\bibitem{zhang2016generating}
Yizhe Zhang, Zhe Gan, and Lawrence Carin.
\newblock Generating text via adversarial training.
\newblock In {\em NIPS workshop on Adversarial Training}, volume~21, 2016.

\bibitem{zhang2017adversarial}
Yizhe Zhang, Zhe Gan, Kai Fan, Zhi Chen, Ricardo Henao, Dinghan Shen, and
  Lawrence Carin.
\newblock Adversarial feature matching for text generation.
\newblock {\em arXiv preprint arXiv:1706.03850}, 2017.

\bibitem{CycleGAN2017}
Jun-Yan Zhu, Taesung Park, Phillip Isola, and Alexei~A Efros.
\newblock Unpaired image-to-image translation using cycle-consistent
  adversarial networks.
\newblock In {\em Computer Vision (ICCV), 2017 IEEE International Conference
  on}, 2017.

\end{thebibliography}
\end{document}